# Artificial Intelligence Framework for Simulating Clinical Decision-Making: A Markov Decision Process Approach

Casey C. Bennett[a,b] and Kris Hauser[b]

[a]*Department of Informatics*
*Centerstone Research Institute*
*44 Vantage Way*
*Suite 280*
*Nashville, TN, USA 37228*

[b]*School of Informatics and Computing*
*Indiana University*
*901 E. 10th. St.*
*Bloomington, IN, USA 47408*

Corresponding Author:
*Casey Bennett*
*Department of Informatics*
*Centerstone Research Institute*
*365 South Park Ridge Road*
*Bloomington, IN 47401*
1.812.336.1156
Casey.Bennett@CenterstoneResearch.org
cabennet@indiana.edu

**Abstract**

Objective: In the modern healthcare system, rapidly expanding costs/complexity, the growing myriad of treatment options, and exploding information streams that often do not effectively reach the front lines hinder the ability to choose optimal treatment decisions over time. The goal in this paper is to develop a general purpose (non-disease-specific) computational/artificial intelligence (AI) framework to address these challenges. This framework serves two potential functions: 1) a simulation environment for exploring various healthcare policies, payment methodologies, etc., and 2) the basis for clinical artificial intelligence – an AI that can "think like a doctor."

Methods: This approach combines Markov decision processes and dynamic decision networks to learn from clinical data and develop complex plans via simulation of alternative sequential decision paths while capturing the sometimes conflicting, sometimes synergistic interactions of various components in the healthcare system. It can operate in partially observable environments (in the case of missing observations or data) by maintaining belief states about patient health status and functions as an online agent that plans and re-plans as actions are performed and new observations are obtained. This framework was evaluated using real patient data from an electronic health record.

Results: The results demonstrate the feasibility of this approach; such an AI framework easily outperforms the current treatment-as-usual (TAU) case-rate/fee-for-service models of healthcare. The cost per unit of outcome change (CPUC) was $189 vs. $497 for AI vs. TAU (where lower is considered optimal) – while at the same time the AI approach could obtain a 30-35% increase in patient outcomes. Tweaking certain AI model parameters could further enhance this advantage, obtaining approximately 50% more improvement (outcome change) for roughly half the costs.

Conclusion: Given careful design and problem formulation, an AI simulation framework can approximate optimal decisions even in complex and uncertain environments. Future work is described that outlines potential lines of research and integration of machine learning algorithms for personalized medicine.

# 1. Introduction

## 1.1 Problem

There are multiple major problems in the functioning and delivery of the modern healthcare system – rapidly expanding costs and complexity, the growing myriad of treatment options, and exploding information streams that often do not, or at most ineffectively, reach the front lines. Even the answer to the basic healthcare question of "What's wrong with this person" often remains elusive in the modern era – let alone clear answers on the most effective treatment for an individual or how we achieve lower costs and greater efficiency. With the expanding use of electronic health records (EHRs) and growth of large public biomedical datasets (e.g. GenBank, caBig), the area is ripe for applications of computational and artificial intelligence (AI) techniques in order to uncover fundamental patterns that can be used to predict optimal treatments, minimize side effects, reduce medical errors/costs, and better integrate research and practice [1].

These challenges represent significant opportunities for improvement. Currently, patients receive correct diagnoses and treatment less than 50% of the time (at first pass) [2]. There is stark evidence of a 13-17 year gap between research and practice in clinical care [3]. This reality suggests that the current methods for moving scientific results into actual clinical practice are lacking. Furthermore, evidence-based treatments derived from such research are often out-of-date by the time they reach widespread use and don't always account for real-world variation that typically impedes effective implementation [4]. At the same time, healthcare costs continue to spiral out-of-control, on pace to reach 30% of gross domestic product by 2050 at current growth rates [5]. Training a human doctor to understand/memorize all the complexity of modern healthcare, even in their specialty domain, is a costly and lengthy process – for instance, training a human surgeon now takes on average 10 years or 10,000 hours of intensive involvement [6].

## 1.2 Goal

The goal in this paper is to develop a general purpose (non-disease-specific) computational/AI framework in an attempt to address these challenges. Such a framework serves two potential functions. First, it provides a simulation environment for understanding and predicting the consequences of various treatment or policy choices. Such simulation modeling can help improve decision-making and the fundamental understanding of the healthcare system and clinical process – its elements, their interactions, and the end result – by playing out numerous potential scenarios in advance. Secondly, such a framework can provide the basis for clinical artificial intelligence that can deliberate in advance, form contingency plans to cope with uncertainty, and adjust to changing information on the fly. In essence, we are attempting to replicate clinician decision-making via simulation. With careful design and problem formulation, we hypothesize that such an AI simulation framework can approximate optimal decisions even in complex and uncertain environments, and approach – and perhaps surpass – human decision-making performance for certain tasks. We test this hypothesis using real patient data from an EHR.

Combining autonomous AI with human clinicians may serve as the most effective long-term path. Let humans do what they do well, and let machines do what they do well. In the end, we may maximize the potential of both. Such technology has the potential to function in multiple roles: enhanced telemedicine services, automated clinician's assistants, and next-generation clinical decision support systems (CDSS) [7,8].

**1.3 Previous work**

In previous work, we have detailed computational approaches for determining optimal treatment decisions at single timepoints via the use of data mining/machine learning techniques. Initial results of such approaches have achieved success rates of near 80% in predicting optimal treatment for individual patients with complex, chronic illness, and hold promise for further improvement [7,9]. Predictive algorithms based on such data-driven models are essentially an individualized form of practice-based evidence drawn from the live population. Another term for this is "personalized medicine."

The ability to adapt specific treatments to fit the characteristics of an individual's disorder transcends the traditional disease model. Prior work in this area has primarily addressed the utility of genetic data to inform individualized care. However, it is likely that the next decade will see the integration of multiple sources of data - genetic, clinical, socio-demographic – to build a more complete profile of the individual, their inherited risks, and the environmental/behavioral factors associated with disorder and the effective treatment thereof [10]. Indeed, we already see the trend of combining clinical and genetic indicators in prediction of cancer prognosis as a way of developing cheaper, more effective prognostic tools [11-13].

Such computational approaches can serve as a component of a larger potential framework for real-time data-driven clinical decision support, or "adaptive decision support." This framework can be integrated into an existing clinical workflow, essentially functioning as a form of artificial intelligence that "lives" within the clinical system, can "learn" over time, and can adapt to the variation seen in the actual real-world population (Figure 1). The approach is two-pronged – both developing new knowledge about effective clinical practices as well as modifying existing knowledge and evidence-based models to fit real-world settings [7,9].

**Figure 1: Clinical decision-making – flow diagram**

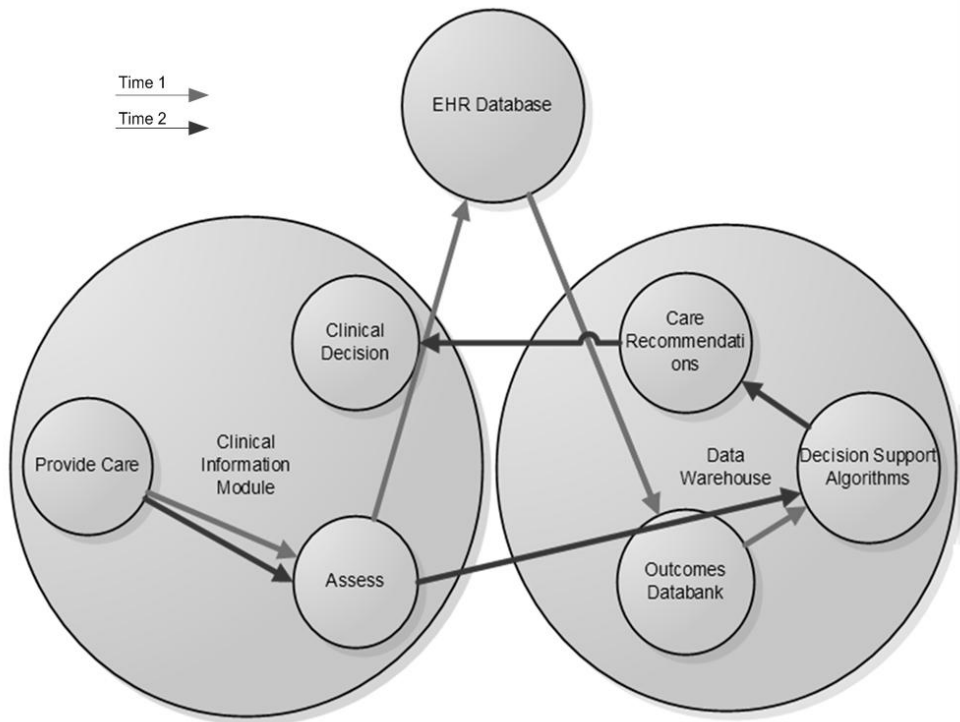

**A high-level example of information flow during the clinical process through an electronic health record (EHR) across time. Information gleaned at Time 1 can be utilized at Time 2. Data flows from the front-end (clinical information module) to a backend EHR database, which can then be pushed into a data warehouse (DW). The DW can analyze this raw data and then push information or recommendations back to the front-end. AI algorithms can "live" in this DW and integrate into an existing clinical workflow, learning over time.**

### 1.4 Computational approaches to sequential decision making

The focus of the current study is to extend the prior work beyond optimizing treatments at single decision points in clinical settings. This paper considers sequential decision processes, in which a sequence of interrelated decisions must be made over time, such as those encountered in the treatment of chronic disorders.

At a broad level, modeling of dynamic sequential decision-making in medicine has a long and varied history. Among these modeling techniques are the Markov-based approaches used here, originally described in terms of medical decision-making by Beck and Pauker [14]. Other approaches utilize dynamic influence diagrams [15] or decision trees [16, 17] to model temporal decisions. An exhaustive review of these approaches is beyond the scope of this article, but a general overview of simulation modeling techniques can be found in Stahl 2008 [17]. In all cases, the goal is to determine optimal sequences of decisions out to some horizon. The treatment of time – whether it is continuous or discrete, and (if the latter) how time units are determined – is a critical aspect in any modeling effort [17], as are the trade-offs between solution quality and solution time [15]. Problems can be either finite-horizon or infinite-horizon. In either case, utilities/ rewards of various decisions can be undiscounted or discounted, where discounting increases the importance of short-term utilities/rewards over long-term ones [18].

Markov decision processes (MDPs) are one efficient technique for determining such optimal sequential decisions (termed a "policy") in dynamic and uncertain environments [18,19], and have been explored in medical decision-making problems in recent years [18,20]. MDPs (and their partially observable cousins) directly address many of the challenges faced in clinical decision-making [17,18]. Clinicians typically determine the course of treatment considering current health status as well as some internal approximation of the outcome of possible future treatment decisions. However, the effect of treatment for a given patient is non-deterministic (i.e. uncertain), and attempting to predict the effects of a series of treatments over time only compounds this uncertainty. A Markov approach provides a principled, efficient method to perform probabilistic inference over time given such non-deterministic action effects. Other complexities (and/or sources of uncertainty) include limited resources, unpredictable patient behavior (e.g., lack of medication adherence), and variable treatment response time. These sources of uncertainty can be directly modeled as probabilistic components in a Markov model [19]. Additionally, the use of outcome deltas, averse to clinical outcomes themselves, can provide a convenient history *meta-variable* for maintaining the central Markov assumption: that the state at time *t* depends only on the information at time *t*-1 [17]. Currently, most treatment decisions in the medical domain are made via ad-hoc or heuristic approaches, but there is a growing body of evidence that such complex treatment decisions are better handled through modeling rather than intuition alone [18,21].

Partially observable Markov decision processes (POMDPs) extend MDPs by maintaining internal *belief states* about patient status, treatment effect, etc., similar to the cognitive planning aspects in a human clinician [22,23]. This is essential for dealing with real-world clinical issues of noisy observations and missing data (e.g. no observation at a given timepoint). By using temporal belief states, POMDPs can account for the probabilistic relationship between observations and underlying health status over time and reason/predict even when observations are missing, while still using existing methods to perform efficient Bayesian inference. MDPs/POMDPs can also be designed as online AI agents – determining an optimal policy at each timepoint (*t*), taking an action based on that optimal policy, then re-determining the optimal policy at the next timepoint (*t*+1) based on new information and/or the observed effects of performed actions [24,25].

A challenge in applying MDP/POMDPs is that they require a data-intensive estimation step to generate reasonable transition models – how belief states evolve over time – and observation models – how unobserved variables affect observed quantities. Large state/decision spaces are also computationally expensive to solve particularly in the partially observable setting, and must adhere to specific Markov assumptions that the current timepoint (t) is dependent only on the previous timepoint (t-1). Careful formulation of the problem and state space is necessary to handle such issues [17,19].

## 1.5 Current work

There have been many applications in other domains, such as robotics, manufacturing, and inventory control [17,19,26]. However, despite such applicability of sequential decision-making techniques like MDPs to medical decision-making, there have been relatively few applications in healthcare [18,19].

Here, we outline a MDP/POMDP simulation framework using agents based on clinical EHR data drawn from real patients in a chronic care setting. We attempt to optimize "clinical utility" in terms of cost-effectiveness of treatment (utilizing both outcomes and costs) while accurately reflecting realistic clinical decision-making. The focus is on the physician's (or physician agent's) optimization of treatment decisions over time. We compare the results of these computational approaches with existing treatment-as-usual approaches to test our primary hypothesis – whether we can construct a viable AI framework from existing techniques that can approach or even surpass human decision-making performance (see Section 1.2).

The framework is structured as a multi-agent system (MAS) for future potential studies, though at the current juncture this aspect is not fully leveraged. However, combining MDPs and MAS opens up many interesting opportunities. For instance, we can model personalized treatment simply by having each patient agent maintain their own individualized transition model (see Discussion). MAS can capture the sometimes synergistic, sometimes conflicting nature of various components of such systems and exhibit emergent, complex behavior from simple interacting agents [14,27]. For instance, a physician may prescribe a medication, but the patient may not adhere to treatment [20].

## 2. Methods
### 2.1 Data

Clinical data, including outcomes, treatment information, demographic information, and other clinical indicators, was obtained from the electronic health record (EHR) at Centerstone for 961 patients who participated in the Client-Directed Outcome-Informed (CDOI) pilot study in 2010 [9], as well as patients who participated in the ongoing evaluation of CDOI post-pilot phase. This sample contained 5,807 patients, primarily consisting of major clinical depression diagnoses, with a significant number of patients (~65%) exhibiting co-occurring chronic physical disorders including hypertension, diabetes, chronic pain, and cardiovascular disease.

Centerstone providers in Tennessee and Indiana see over 75,000 distinct patients a year across over 130 outpatient clinical sites. Centerstone has a fully-functional system-wide EHR that maintains virtually all relevant patient records.

In all simulations, 500 randomly selected patients were used. All other aspects, probabilities, and parameters for modeling were estimated directly from the EHR data (e.g. average cost per service, expected values of outcome improvement and deterioration, and transition model probabilities).

In all subsequent simulations, a single physician agent with a caseload of 500 randomly selected patients was used. The framework can handle multiple physician agents (from a programming code standpoint, see Section 2.2), but at the present time there are no variable behaviors across physicians (e.g. variable attitudes towards outcome significance, where one physician pays a lot of attention to outcome observations and another does not). The physician agent must make a treatment decision for each patient at each timepoint over the course of seven sessions (plus baseline/intake, max total sessions=8).

The primary outcome of interest used here is the Outcome Rating Scale (ORS) component of the CDOI assessment, which is a validated ultra-brief measure of functioning and symptomology for chronic and acute mental disorders, with over a decade of research supporting its use in similar settings and populations [28]. The ORS correlates highly with lengthier, traditional outcome measures such as OQ-45 and the Quality of Life Scale (QOLS) [29]. The ORS has been shown previously to be useful as the basis for machine learning algorithms to predict individualized patient treatment response [9].

The utility metric, which is used to evaluate the quality of decisions in a model, is cost per unit change (CPUC), which measures the cost in dollars it takes to obtain one unit of outcome change (delta) on a given outcome [30]. In essence, CPUC is a *relative* measure of cost-effectiveness of treatment, given as a ratio between costs and outcome change. In this study, CPUC was calculated using the change in CDOI-ORS over time (delta) - the delta calculation varying dependent on the formulation of the transition model (Section 2.4). However, CPUC could be calculated for any disease and/or outcome measure – e.g. blood pressure, cancer metastasis stage, hospitalization days, quality-adjusted life years (QALYs). Hence, the use of CPUC, rather than directly using outcomes, for utility/rewards in the modeling framework is principle to keeping the model general purpose (non-disease-specific).

**2.2 Framework overview**

A general framework overview can be seen in Figure 2, which is further elaborated in Sections 2.3-2.5 below. The agents (shown in double-line borders) encapsulate the characteristics and functions of their respective real-life counterparts – e.g. patient agents incorporate individual patient-specific data and individualized transition models (see Discussion) while physician agents maintain beliefs about patients' health status and treatment effects and have decision-making capabilities.

**Figure 2: Framework overview**

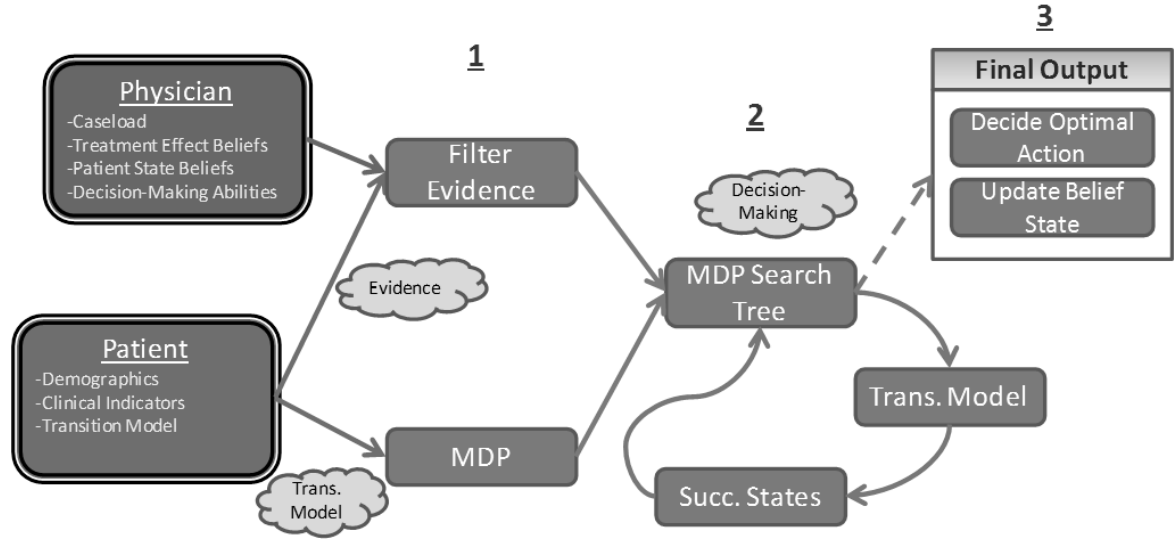

The types of agents are shown in double-line borders. The other boxes represent various aspects of the model. The general flow is: 1) create patient-specific MDPs/physician agent filters evidence into existing beliefs, 2) recurse through MDP search tree to determine optimal action, and 3) perform treatment action and update belief states

Figure 2 also displays the general algorithm (implemented in Python 2.7, www.python.org), where initially patient-specific MDPs are created from the transition models and physician agents must incorporate patient-specific evidence/information into existing beliefs prior to decision-making at each timepoint. The decision-making process then recurses down the MDP search tree, resulting finally in a determination of an optimal current action and updates to patient belief states.

The algorithm steps are as follows:
1) Create patient and physician agents
2) Create patient-specific MDP

*Then, for each timepoint (if not horizon):*

3) Calculate current outcome delta, physician agent filters evidence
4) Determine optimal current action via MDP search tree
5) Perform action and update belief states
6) If action ≠ not treat, return to step 3

## 2.3 POMDP decision-making environment

We model the decision-making environment as a finite-horizon, undiscounted, sequential decision-making process in which the *state $s_t$* from the *state space S* consists of a patient's health status at time *t*. At each time step the physician agent makes a decision to treat or stop treatment (an *action $a_t$* from the binary *action space A={0,1}*). Here time corresponds to the number of treatment sessions since the patient's first visit (typically one session=one week). The physician agent receives rewards/utilities, and is asked to pick actions in order to maximize overall utilities. Similar decision-making models were used in [19,31,32]. We can model this decision as a dynamic decision network (DDN, a type of dynamic Bayesian network), as seen in Figure 3.

**Figure 3: Dynamic decision network (DDN)**

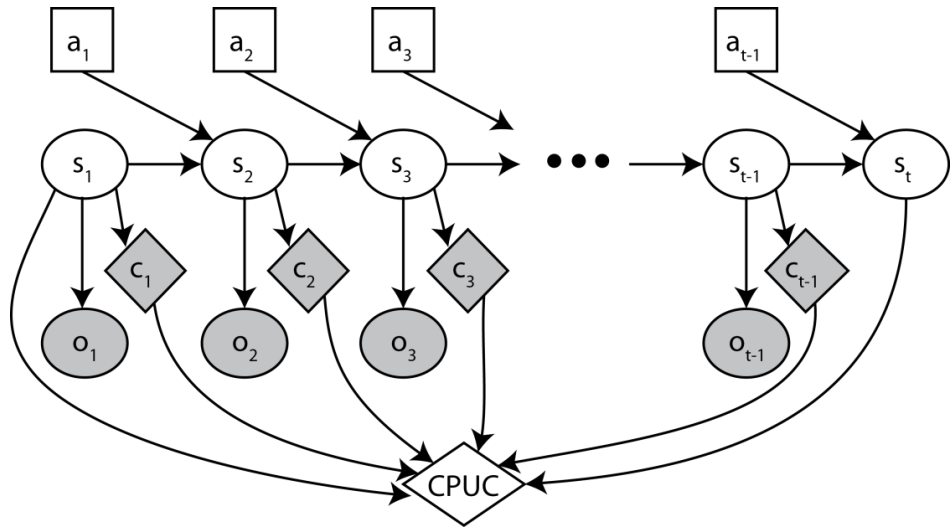

A dynamic decision network for clinical decision-making. a = action (e.g. treatment option), s = state (patient's actual underlying health status), o = observation (patient's observed status/outcome), c = treatment costs, CPUC = utilities/rewards (cost per unit change of some outcome measure). The subscripts represent time slices (e.g. treatment sessions).

Here:
- *s* = State (patient's actual status, not directly observable)
- *o* = Observation (patient's observed status/outcome, can be missing)
- *a* = Action (e.g. treatment option, not treat)
- *c* = Treatment costs
- CPUC = Utilities/Rewards (in this case, patient's CPUC)

We are interested in the following: at time *t*, what is the optimal action (*a*) to choose? The answer depends on what the agent has observed so far, the beliefs the agent has inferred from those observations, and what the agent expects might happen in the future.

In our current implementation, health status is taken to be the delta value (change over time, see Section 2.4) of the CDOI-ORS (ΔCDOI), given an observation at time *t*, but more generally state models could represent underlying characteristics of disease, clinical/demographic characteristics, and individual genetic predispositions. *States* – actual, $s_t$, and belief states, $b_t$ (see below) – are derived by discretizing continuous CDOI delta values into five discrete bins: High Deterioration (ΔCDOI < −4), Low Deterioration (−4 <= ΔCDOI < −1), Flatline/Stable (−1 <= ΔCDOI <= 1), Low Improvement (1 < ΔCDOI <= 4) and High Improvement (ΔCDOI > 4). These bins are derived from research-validated clinical categorization of significant outcome delta classes for CDOI-ORS, described by Miller et al. [28]. This binning is necessary for computational tractability (Section 2.5.1).

We model the effects of actions on the state using a *transition model* (*TR*) that encodes the probabilistic effects of various treatment actions:

$$TR(s_{t+1}, s_t, a_t) = P(s_{t+1} \mid s_t, a_t) \tag{1}$$

which is assumed stationary (invariant to the time step).

The physician agent's performance <u>objective is to maximize the improvement in patient health, as measured by the change in CDOI-ORS score at the end of treatment, while minimizing cost of treatment (e.g. by stopping treatment when the probability of further improvement is low)</u>. These two competing objectives are combined via the cost per unit change (CPUC) metric that incorporates both outcome delta (change over time) and treatment costs. Treatment strategies with low CPUC are cost effective (high utility). An optimal strategy $\pi^*$ that *minimizes* CPUC is given as:

$$\pi^* = \arg\min_{\pi} CPUC(\pi) = \arg\min_{\pi} E \begin{cases} \dfrac{C_T(\pi)}{\Delta CDOI(\pi)}, & \text{if } \Delta CDOI(\pi) \geq 1 \\ \dfrac{C_T(\pi)}{1} + (1 - \Delta CDOI(\pi)) * CPS, & \text{else} \end{cases} \quad (2)$$

where CPS=cost per service for a given session, $C_T$ is a random variable denoting accumulated cost at the end of treatment, $E$ indicates expected value (since we are calculating over future events), and $\Delta CDOI(\pi)$ is calculated as:

$$\Delta CDOI(\pi) = CDOI_T(\pi) - CDOI_0 \quad (3)$$

where $CDOI_T$ is a random variable denoting the CDOI-ORS value at the end of treatment. The expectation is taken over future patient health trajectories up to a finite horizon $T$, here taken to be $T$=8 treatments (based on the typical average number of sessions amongst Centerstone's outpatient population). In cases where the CDOI delta $\leq 0$, we rescale the delta values so that as a utility metric, given equal costs, delta=0 is effectively one unit worse than delta=1, delta= −1 is one unit worse than delta=0, and so on. This is done by adding additional costs to the CPUC for delta=1.

In all cases, the treatment decision must be based on the belief state rather than the true underlying state (which cannot be directly observed) – that is, a strategy $\pi$ is defined as a map from belief states to actions: $\pi : 2^S \rightarrow A$. In other words, the physician agent's reasoning is performed in a space of *belief states*, which are *probability distributions* over the patient's health status, $b_t(s)=P(s_t=s)$. For instance, we cannot directly observe a patient's disease state (e.g. diabetes); rather, we take measurements of symptoms (e.g. blood glucose) and attempt to classify the patient into some underlying disease or health state. Furthermore, in approximately 30% of our data, the clinician makes a treatment decision when the CDOI-ORS observation is missing (i.e. partially observable environment), and the belief state must be inferred from previous belief states (see below). The belief state categories are the same as those described above for the true underlying state (High Deterioration, Flatline, etc.) In future work, we would like to extend the system to reason optimally when integrating unobserved health factors based on their probabilistic relationship to observed clinical/demographic characteristics, as well as account for non-deterministic effects of variable treatment options (see Discussion).

Unconditional on observations, the next belief state, $b_{t+1}$, is predicted from the prior belief (over all possible prior states) using the following exact equation:

$$b_{t+1}(s_{t+1}) = Predict(b_t, a_t) = \sum_{s_t} TR(s_{t+1}, s_t, a_t) b_t(s_t) \quad (4)$$

By repeated application of this predict operation our system can compute forecasts of the patient's health status into the future.

Over time, uncertainty of the belief state is reduced by relating the *observations* that are actually seen by the physician to the health status; in other words, when observations are available, we utilize them to update the belief states. In our problem, observations $o_t$ are drawn from the *observation space* O = {missing} ∪ CDOI. In the case of a missing observation, $o_t$ = {missing}, the belief $b_t(s)$ is maintained as is after the observation. This provides a probabilistic *observation model*, which defines the relationship between the true underlying state and possible observations:

$$O(o_t, s_t) = P(o_t | s_t) \quad (5)$$

Upon receiving an observation $o_t$, we find the posterior belief over the patient's state using the update operation, which uses Bayes rule to derive the backward relationship from an observation to the state distribution:

$$b_{t|o_t}(s) = Update(b_t, o_t) = \frac{1}{Z} * P(o_t | s) \, b_t(s) \quad (6)$$

where Z is a normalization factor equal to the unconditional probability of observing $o_t$. For all patients and their current belief state, the physician agent maintains a continuous-valued estimate of CDOI-ORS at the given timepoint by applying a Gaussian model of average treatment effect (estimated from the EHR data) to the CDOI-ORS value at the previous timepoint given the predicted belief state, $b_t(s)$ (e.g. High Improvement). This continuous-valued CDOI-ORS belief can be recalculated as a delta and then re-binned into states for future prediction steps, $b_{t+1}(s)$.

## 2.4 Estimation of transition models

The goal of the transition model is to use the history of health status – i.e. outcome delta – to predict the probability distribution of future health states on the subsequent time step. Let $h_t$ denote the history of observations $(o_1,\ldots,o_t)$. We compare three model classes for predicting the change in CDOI-ORS from the current to the next step: $\Delta CDOI_t = CDOI_{t+1} - CDOI_t$.

1) **$0^{th}$ order** – a raw stationary distribution over $\Delta CDOI_t$ independent of history (i.e. the probability of treatment effect, regardless of improvement/deterioration seen thus far): $P(\Delta CDOI_t \mid h_t) = P(\Delta CDOI_t)$
2) **$1^{st}$ order autoregressive (Local)** – the distribution over $\Delta CDOI_t$ depends only on change since the previous timepoint (local change): $P(\Delta CDOI_t \mid h_t) = P(\Delta CDOI_t \mid \Delta CDOI_{t-1})$.
3) **Global average** – the distribution over $\Delta CDOI_t$ depends on the *entire* patient history (i.e. delta since baseline): $P(\Delta CDOI_t \mid h_t) = P(\Delta CDOI_t \mid (CDOI_t - CDOI_0))$.

The $0^{th}$ order model ignores any effect of the history on future patient outcomes and treats each patient like a new average patient regardless of previous observations. For instance, even if the patient has already experienced significant outcome improvement, the $0^{th}$ order model still assumes they are just as likely to improve in the future. On the other hand, the $1^{st}$ order local model uses the short-term trajectory of prior improvement/deterioration in order to gain somewhat better forecasting ability, but only since the most recent timepoint (previous treatment session). One potential drawback of this method is that it may be fooled by large and/or spurious short-term oscillations in patient outcomes. The global averaging technique looks at trends over a longer time horizon (change since baseline, t=0). It provides the most comprehensive measure but is less sensitive to recent changes, which can be significant for real-world treatment decisions. The global technique also maintains the Markov property required for MDP use by capturing the total history of outcome change as a state meta-variable of a given timepoint (see Section 2.5.1). In future work, the potential also exists to combine of models of differing orders to improve forecasting (e.g. $1^{st}$ order/local and global deltas).

For each transition model class, we build a discrete conditional probability table over the values of the independent variable using observed statistics from our EHR data (in other words, using a separate sample of patients from the EHR, we estimate the transition probabilities needed for the model). To obtain sufficient sample size for the estimation procedure, we bin deltas into 5 bins (High Deterioration, Low Deterioration, Flatline/Stable, Low improvement, High Improvement) based on research-validated clinical categories (see Section 2.3) [28]. For each model, we estimated transition probabilities using maximum likelihood from EHR data. As typical for many EHR systems, our dataset only contains CDOI-ORS data for patients to the point of treatment termination – i.e., we have no information on patients' health status after the discontinuation of services at Centerstone. As noted elsewhere, the collection of such "natural history" disease data is fraught with many challenges, ethical and otherwise, particularly if such untreated conditions pose significant health risks [18]. Hence, we make the coarse approximation that untreated individuals, on average, remain roughly constant at $\Delta CDOI_t=0$.

### 2.5 Decision-making strategies
### 2.5.1 MDP models

To determine optimal actions via the DDN, we compute an *optimal* treatment strategy via exploration of a belief-space search tree (MDP search tree). Here, we present an online approach where the system continually plans, executes, observes, and re-plans the optimal treatment strategy from any given timepoint [33].

The MDP/POMDP search tree, also sometimes referred to as a stochastic decision tree [34], explores the beliefs (*b*) obtained for all possible actions (*a*) and probabilistic treatment outcomes (observations, *o*) out into the future (Figure 4). The tree alternates layers of decision nodes – in which the physician agent has a choice of action – and chance nodes – in which the environment (including other agents' actions, non-deterministic treatment effects, and so on) yields uncertain changes to the patient health status. For stop-treatment actions, we also construct *terminal* nodes. In our case, we end enumerating layers at the finite horizon, T=8 (producing a tree of depth 16 in total).

We then make a backwards pass to compute the optimal decisions at each decision node that optimizes CPUC. The first step in this pass computes the overall CPUC at leaves and terminal nodes. Then, we recursively backup the optimal CPUC for interior nodes all the way to the root. The backup operation for chance nodes calculates the expected CPUC over its children (based on probabilities), while at decision nodes the backup picks the action that leads to the optimal (in this case, minimal) CPUC for its subtree. Once this operation is complete, we can compute the optimal strategy. The optimal strategy is a *subtree* obtained by keeping all children of chance nodes and the single child corresponding to the optimal action at each decision node. This optimal strategy represents a treatment *plan* given current information.

**Figure 4: MDP search tree**

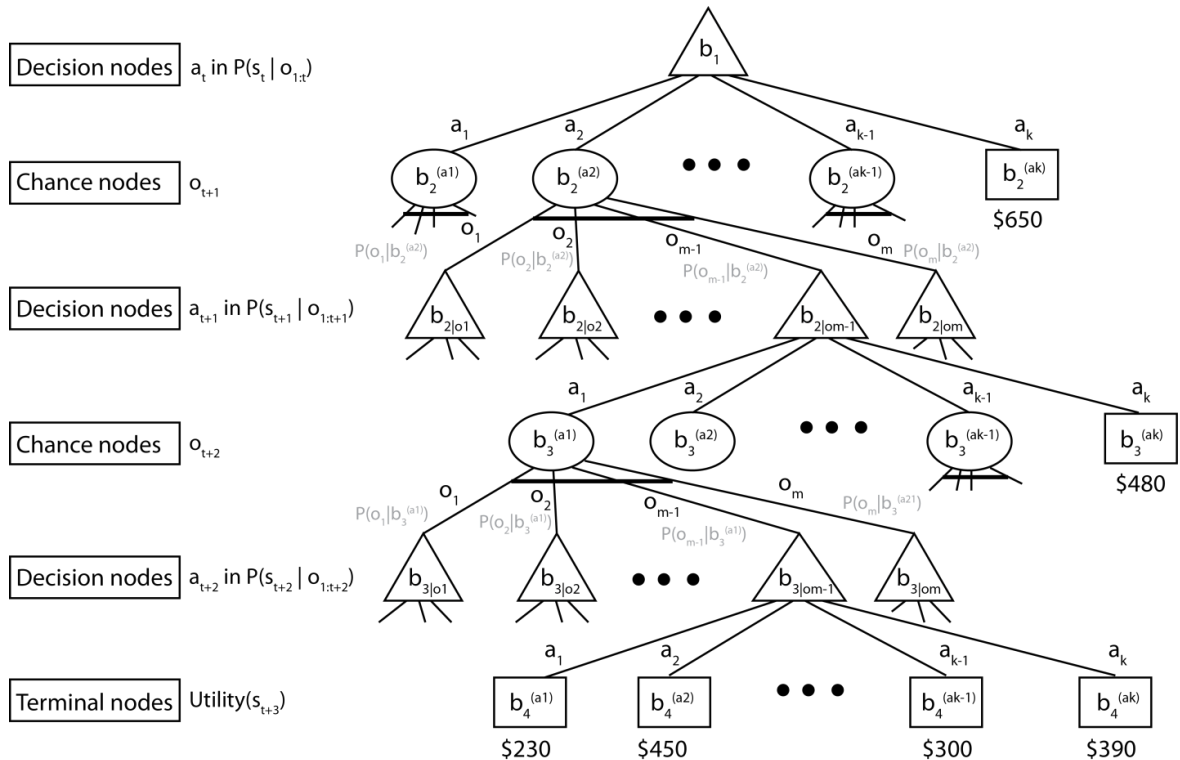

An example MDP search tree for clinical decision-making. At each layer/time slice, we have a series of decision and chance (or terminal) nodes. Decision nodes represent choices of actions ($a$, given $k$ possible actions), while chance nodes represent possible resulting observations ($o$, given $m$ possible observations). Terminal nodes present a utility measure (CPUC, represented in dollars). As none of these actions or observations has occurred yet, we must use probabilistic beliefs ($b$) at each planning step.

Where the various branches, $b$, can relate to possible observations, $o=\{o_1...o_m\}$, at the decision nodes and actions, $a=\{a_1...a_k\}$, at the probabilistic nodes. Each node contains some utility value (here a CPUC value shown in dollars).

A primary challenge in this approach is exploring the large search tree that can result. Here, we consider 5 observation bins and a "stop treatment" action with no subsequent branches, resulting in a search tree with branching factor = 5 and depth = 10 and the generation of over 100,000 belief states in the worst case. While tractable for the problem at hand, more complex decision problems – such as larger state spaces, action types, or time frames – may require approximate solutions [15].

**2.5.2 Comparative models**

For comparative purposes with the MDP models, we evaluated several simpler decision-making approaches. These alternative strategies provide context for interpreting the optimality of the MDP results.

We considered two heuristic policies that represent current healthcare models. The MDP models were constructed only with the $1^{st}$ order and global transition models described in section 2.4, as the MDP policy is trivial in the $0^{th}$ order case because ΔCDOI from *t* to *t+1* is assumed independent of prior history. As such, the $0^{th}$ order models essentially represent their own decision-making approach – henceforth referred to as **Raw Effect** models, given they consider the raw effect of treatment without consideration of outcome change history. Given its optimistic nature, the Raw Effect model results in always treating till the horizon. The Raw Effect model provides an *upper baseline*, which represents the overly-optimistic scenario of assuming treatment always results in patient improvement. This is, in effect, an approximation of the <u>fee-for-service model</u> prevalent in U.S healthcare. We also considered a **Hard Stop** policy after the third treatment session. The Hard Stop strategy provides a *lower baseline*, simulating the worst-case scenario of simply stopping treatment after some arbitrary timepoint without consideration for outcomes (minimizing costs). This is, in effect, an approximation of the <u>case-rate/capacitated model</u> used by many insurance companies.

Additionally, for the $1^{st}$ order and global transition models, we consider two simpler decision-making approaches (that do consider outcome change history):

1) **Max Improve** that assumes the treatment effect of maximum probability always occurs for a given action. This could be considered a "winner-take-all" approach.
2) **Probabilistic** models select an action at random, where the action is chosen with probability proportional to its likelihood to improve CDOI. For example, if treatment was predicted to improve CDOI with probability 0.9, then the strategy would flip a biased coin with probability 0.9 to decide whether to treat.

Both of the above models – as well as the Raw Effect model – only consider the probabilities of treatment effects for a given action from *t* to *t+1*, unlike the MDP approach which can calculate utilities of a *current action* across multiple future timepoints considering possible action sequences, treatment effects, and contingency plans. We consider them here to evaluate whether the added complexity of the MDP models results in justifiable performance improvement in a healthcare setting.

**2.6 Simulation analysis approach**

For purposes of analysis, simulation experiments were performed across multiple permutations of the transition and decision-making models laid out in Sections 2.4 and 2.5 (henceforth termed *constructs*). We also considered datasets with and without missing observation points.

Additionally, for the MDP model solved via DDN/search tree, the significance of the outcomes was varied via an *outcome scaling factor* (OSF) that adds in scaled outcome values {0-1} as an additional component of the utility metric. The outcomes (current *delta*) are scaled and flipped based on the maximum possible delta for a patient at a given timepoint (*delta$_{max}$*), so that higher values (near 1) are worse than lower values (given that we are attempting to minimize CPUC):

$$CPUC_{Final} = CPUC + OSF * \left(\frac{delta_{max} - delta}{delta_{max}}\right) \qquad (7)$$

Although any scaling to the range *y* = {0-1} and flipped so that OSF = 1− *y* would work. Increasing this OSF above 0 increases the added importance of outcomes in the decision-making process. It should be noted – this is "added" influence, because outcomes are already accounted for in the basic CPUC reward/utility calculation, even when this factor is set to 0. When set to 0, outcomes are considered equally important as costs.

For the probabilistic decision-making models (Section 2.5.2), it was necessary to perform multiple runs (*n*=10) of each construct in order to build a statistical sample from which to derive mean values for CPUC, outcomes, etc. given these probabilistic models are purely non-deterministic nature. For the other decision-making models, this was unnecessary.

## 3. Results
### 3.1 General results

Nearly 100 different constructs were evaluated during this study. For brevity, a sampling of the main results is shown in Table 1 (with OSF=0). In general, the purely probabilistic decision-making models performed poorly, and are not shown here. In all tables, results are based averages/percentages across all patients (n=500, see Section 2.1) in each construct simulation. <u>As defined in Section 2.3, the goal here (i.e. optimality) is defined as maximizing patient improvement while minimizing treatment costs, which equates to minimizing CPUC</u>. Additionally, we would prefer models that maintain reasonably high average final delta values and lower standard deviations of delta values.

Table 1 shows the results of the Hard Stop (i.e. case rate) and the Raw Effect (i.e. fee-for-service) decision models first. It should be noted that – for the former – CPUC is reduced but still generally higher than other decision making approaches while outcomes are very low, and that – for the latter – outcomes are improved but still generally lower than other decision making approaches while CPUC is significantly higher. In short, neither is optimal.

**Table 1: Model simulation results**

| Decision Model | Transition Model | Missing Obs[1] | CPUC[2] | Avg Final Delta[3] | Std Dev Final Delta[3] | Avg # of Services | % Patients Max Dosage[4] |
|---|---|---|---|---|---|---|---|
| Hard Stop | N/A | No | 262.30 | 3.22 | 7.80 | 3.00 | 0% |
| Hard Stop | N/A | Yes | 305.53 | 2.56 | 8.07 | 3.00 | 0% |
| Raw Effect | 0th Order | No | 470.33 | 4.69 | 8.66 | 8.00 | 100% |
| Raw Effect | 0th Order | Yes | 497.00 | 4.73 | 8.45 | 8.00 | 100% |
| Max Improve | 1st Order | No | 297.47 | 6.24 | 7.87 | 5.35 | 30% |
| Max Improve | 1st Order | Yes | 303.85 | 5.77 | 8.25 | 5.32 | 29% |
| MDP | 1st Order | No | 228.91 | 5.85 | 7.12 | 4.11 | 4% |
| MDP | 1st Order | Yes | 237.21 | 5.11 | 7.37 | 4.03 | 3% |
| Max Improve | Global | No | 256.44 | 6.41 | 6.92 | 4.79 | 24% |
| Max Improve | Global | Yes | 251.83 | 6.07 | 6.90 | 4.76 | 20% |
| MDP | Global | No | 181.72 | 6.07 | 6.42 | 4.23 | 11% |
| MDP | Global | Yes | 189.93 | 5.59 | 6.44 | 4.11 | 9% |

[1] Missing observation
[2] Cost per Unit Change
[3] Final Delta = change in outcome from baseline to end of treatment
[4] Percent of patients receiving maximum number of treatment sessions

More sophisticated decision-making models – including the Max Improve and the MDP models – performed much more optimally across various constructs. This included constructs with the inclusion of missing observations, which is a realistic challenge faced by any healthcare decision-maker. Generally speaking, the MDP decision-making models outperformed the Max Improve models in terms of minimizing CPUC, which was the primary metric of interest. However, the MDP approach generally achieved slightly lower outcome deltas, given an OSF=0 (see section 3.2). The MDP models did have slightly lower standard deviations for outcome deltas across patients except for a few constructs. In general, the MDP models appeared to be more consistent in terms of our definition of optimality. We can also see that performance increases as we move from the $1^{st}$ order to global transition models.

These more sophisticated AI decision-making approaches are superior to the aforementioned, more simplistic methods that are commonly employed in many healthcare payment methodologies (case rate, fee-for-service). <u>An MDP model using a global transition model obtained higher outcomes than the Raw Effect (i.e. fee-for-service) model at significantly lower CPUC ($189 vs. $497), even in the face of missing observations</u>.

It should also be noted that the same values in Table 1 were calculated for the broader patient population from Centerstone's EHR (Centerstone currently operates in a fee-for-service payment model). These "treatment-as-usual" averages were estimated as: CPUC ≈ $540, final CDOI-ORS delta ≈ 4.4 ± 9.5, and number of services ≈ 7.1. These real-world values roughly approximate the simulated values for the Raw Effect decision-making model, providing some ground validity for the simulation approach.

**3.2 MDP variation in outcome scaling factor**

Since the MDP decision-making approaches consider utilities inherently as part of their decision-making approach, they provide additional opportunities for model refinement over other approaches, such as the simpler Max Improve model (Section 3.1). One way to refine the model is by adjusting the outcome scaling factor (OSF, Section 2.6). The results of such adjustment for the MDP model (using the global transition model and inclusion of missing observations) are shown in Table 2. Additionally, we ran the same experiment on a variation of the MDP model which takes the action with maximum probability (MaxProb) at each chance node, rather than the average/expected values (see Section 2.5.1). This is shown in Table 3.

**Table 2: Outcome scaling – normal MDP**

| Outcome Scaling Factor | Decision Model | Transition Model | Missing Obs[1] | CPUC[2] | Avg Final Delta[3] | Std Dev Final Delta[3] | Avg # of Services | % Patients Max Dosage[4] |
|---|---|---|---|---|---|---|---|---|
| 0 | MDP | Global | Yes | 189.93 | 5.59 | 6.44 | 4.112 | 9% |
| 1 | MDP | Global | Yes | 216.54 | 5.71 | 6.58 | 4.32 | 12% |
| 2 | MDP | Global | Yes | 219.95 | 6.17 | 6.58 | 4.62 | 14% |
| 3 | MDP | Global | Yes | 223.74 | 6.24 | 6.63 | 4.68 | 16% |
| 4 | MDP | Global | Yes | 235.87 | 6.26 | 6.62 | 4.70 | 16% |
| 5 | MDP | Global | Yes | 236.99 | 6.32 | 6.63 | 4.76 | 18% |
| 6 | MDP | Global | Yes | 243.87 | 6.29 | 6.80 | 4.80 | 19% |
| 10 | MDP | Global | Yes | 249.03 | 6.35 | 6.90 | 4.92 | 21% |

[1]Missing observation
[2]Cost per unit change
[3]Final Delta = change in outcome from baseline to end of treatment
[4]Percent of patients receiving maximum number of treatment sessions

**Table 3: Outcome scaling – MaxProb MDP**

| Outcome Scaling Factor | Decision Model | Transition Model | Missing Obs[1] | CPUC[2] | Avg Final Delta[3] | Std Dev Final Delta[3] | Avg # of Services | % Patients Max Dosage[4] |
|---|---|---|---|---|---|---|---|---|
| 0 | MDP | Global | Yes | 189.93 | 5.59 | 6.44 | 4.112 | 9% |
| 1 | MDP | Global | Yes | 216.54 | 5.71 | 6.58 | 4.32 | 12% |
| 2 | MDP | Global | Yes | 219.95 | 6.17 | 6.58 | 4.62 | 14% |
| 3 | MDP | Global | Yes | 223.74 | 6.24 | 6.63 | 4.68 | 16% |
| 4 | MDP | Global | Yes | 235.87 | 6.26 | 6.62 | 4.70 | 16% |
| 5 | MDP | Global | Yes | 236.99 | 6.32 | 6.63 | 4.76 | 18% |
| 6 | MDP | Global | Yes | 243.87 | 6.29 | 6.80 | 4.80 | 19% |
| 10 | MDP | Global | Yes | 249.03 | 6.35 | 6.90 | 4.92 | 21% |

[1]Missing observation
[2]Cost per unit change
[3]Final Delta = change in outcome from baseline to end of treatment
[4]Percent of patients receiving maximum number of treatment sessions

In Table 2, we can see that adjusting the OSF from 0 to 10 results in increasing outcome deltas, as well as increasing CPUC's. For the normal MDP, we appear to reach some sort of maxima in outcomes around OSF=3 or 4 (which is reflected in the MaxProb MDP below). After this, CPUC continues to rise, but outcome deltas show minimal improvement, even at OSF=10 and beyond. Effects on CPUC and outcome delta for both models using OSF ranging [0,15] can be seen in Figure 5.

We can see in table 3 that the MaxProb MDP show little change across OSF values in terms of CPUC or outcome deltas. This is probably evidence of some global maxima for CPUC ≈ $220-225 and outcome delta ≈ 6.25-6.35. Existence of such maxima should be expected given the optimization-problem nature of the current framework. It does suggest that such problems may be ripe for application of other optimization techniques for determining model parameters such as OSF or action/treatment decision thresholds.

It should be noted that with OSF=3, the normal MDP obtains CPUC=$224 and outcome delta=6.24 in 4.68 treatment sessions. This appears to be an optimal decision-making model across all model formulations given both CPUC and outcome deltas. It also vastly outperforms the aforementioned treatment-as-usual case rate/fee-for-service models (see Section 3.1). <u>Essentially, we can obtain approximately 50% more improvement (outcome change) for roughly half the costs.</u>

**Figure 5: Outcome scaling effects**

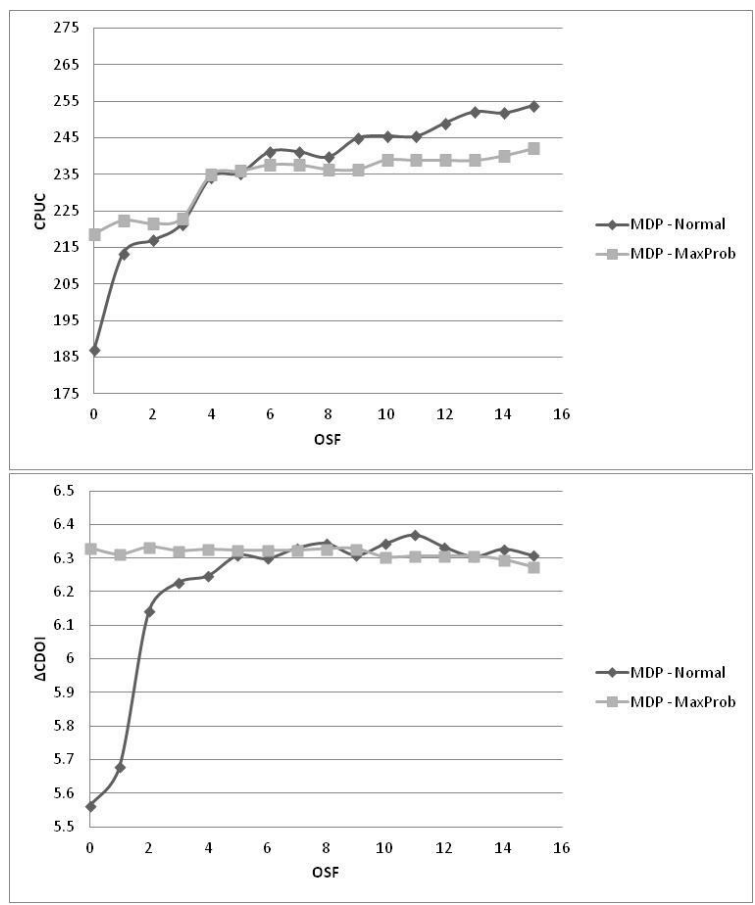

**The figure shows the effect of different values of the OSF parameter on CPUC (top) and outcome delta, i.e. ΔCDOI (bottom), for both the MDP-Normal and MDP-MaxProb Models. The delta values level off around ≈ 6.25-6.35 while CPUC values continue to increase.**

## 4. Discussion
### 4.1 Summarization of findings

The goal in this paper was to develop a general purpose (non-disease-specific) computational/AI framework in attempt to address fundamental healthcare challenges – rising costs, sub-optimal quality, difficulty moving research evidence into practice, among others. This framework serves two potential purposes:

1) A simulation environment for exploring various healthcare policies, payment methodologies, etc.
2) The basis for clinical artificial intelligence – an AI that can "think like a doctor"

This framework utilizes modern computational approaches to learn from clinical data and develop complex plans via simulation of numerous, alternative sequential decision paths. It determines optimal actions based on what has been observed so far, the beliefs inferred from those observations, and what we expect might happen in the future. It must do so in a dynamic environment as an online agent – which continually plans, observes, and re-plans over time as evidence/facts change. The framework is structured as a multi-agent system, which in future work has potential to account for the various conflicting and synergistic interactions of various components in the healthcare system.

The results shown here demonstrate the feasibility of such an approach relative to human decision-making performance. Even in its early stages of development, such an AI framework easily outperforms the current treatment-as-usual case-rate/fee-for-service models of healthcare (Section 3.1). Sophisticated modeling approaches utilizing MDPs and DDNs can be further tweaked to provide finer control over the utility input of higher outcomes or lower costs (Section 3.2), providing robust tools for simulation and AI development in the medical domain.

### 4.2 Future work

Despite the work presented here, there are many areas ripe for exploration in this domain. We are particularly interested in:

- **Personalized transition models** – Integrate machine learning algorithms for optimal treatment selection for each patient.
- **Natural history of disease progression** – For patients who receive no treatment as an action, consider natural history rather than assuming they remain stable.
- **Variable physician agents** – Vary behavior/decision-making by physician depending on individual physician priors (e.g., variable perceptions of outcome significance).
- **Variable patient agents –** Consider variable patient behaviors (e.g. non-adherence to treatment, not taking medications, missing appointments).
- **Optimization methods for thresholds** – To determine cutoffs of decision points between actions, rather than a priori determination via statistical methods. Agent would determine such cut-points "on the fly."
- **Gaussian noise for activity effects** – More realistic modeling of transitions between health states.
- **Improved non-deterministic choices** – For example, Monte Carlo simulations, rather than simply picking the max probability or random number, to determine if this yields better results.
- **Better state conceptualization** – Consider combined global/local ($1^{st}$ order) delta state space to capture both overall patient progress as well as short-term trajectories.
- **Better utility conceptualization** – Consider alternative ways to estimate utility of a given action (e.g. different methods for weighting costs and outcomes).

For example, given the multi-agent design, the system can be modeled on an individual, personalized treatment basis (including genetics), i.e. "personalized medicine" [10]. In previous work (see Introduction), we have described using machine learning methods to determine optimal treatments at a single timepoint for individual patients [7,9]. Such methods could be combined into the sequential decision AI framework described here, simply by incorporating the output probabilities of those single-decision point treatment models into the transition models used by the sequential decision-making approaches [35]. As such, each patient agent could maintain their own individualized transition model, which could then be passed into the physician agent at the time of decision-making for each patient. This is a significant advantage over a "one-size-fits-all" approach to healthcare, both in terms of quality as well as efficiency [36,37].

There are, of course, potential ethical issues about how we might use such quality/performance information as the basis for clinician reimbursement and/or clinical decision-making (e.g. pay-for-performance), but these are broader issues that transcend whether we use artificial intelligence techniques or not [38].

Other opportunities include considering combinations of global/local (1$^{st}$ order) deltas to capture both long-term prognosis as well as short-term trajectories, modeling variable patient behaviors (e.g. non-adherence to medications), testing alternative utility metrics, and utilizing optimization techniques to determine decision cutoff thresholds between actions on-the-fly.

**4.3 Conclusion**

At the end of the day, if we can predict the likely result of a sequence of actions/treatment for some time out into the future, we can use that to determine the optimal action *right now*. As recently pointed out by the Institute of Medicine, does it make sense to continue to have human clinicians attempt to estimate the probabilistic effects of multiple actions over time, across multitudes of treatment options and variable patient characteristics, in order to derive some intuition of the optimal course of action? Or would we be better served to free them to focus on delivery of actual patient care [39]? The work presented here adds to a growing body of evidence that such complex treatment decisions may be better handled through modeling than intuition alone [18,21]. Furthermore, the potential exists to extend this framework as a technical infrastructure for delivering personalized medicine. Such an approach presents real opportunities to address the fundamental healthcare challenges of our time, and may serve a critical role in advancing human performance as well.

**Acknowledgements**

This research is funded by the Ayers Foundation, the Joe C. Davis Foundation, and Indiana University. The funders had no role in the design, implementation, or analysis of this research. The author would like to acknowledge the support of the following Centerstone Research Institute staff in this work: Dr. Tom Doub, Dr. Dennis Morrison, Dr. April Bragg, and Dr. Rebecca Selove.  The opinions expressed herein do not necessarily reflect the views of Centerstone, Indiana University, or their affiliates.

**Conflict of interest**

The authors have no conflict of interest related to the research presented herein.

**References**

[1]   V.L. Patel, E.H. Shortliffe, M. Stefanelli, P. Szolovits, M.R. Berthold, R. Bellazzi, et al., The Coming of Age of Artificial Intelligence in Medicine, *Artif Intell Med*, (2009) 46(1): 5–17.

[2]   E.A. McGlynn, S.M. Asch, J. Adams, J. Keesey, J. Hicks, A. DeCristofaro A, et al., The quality of health care delivered to adults in the United States, *N Engl J Med*, (2003) 348(26): 2635-2645

[3]   M.S. Bauer, A review of quantitative studies of adherence to mental health clinical practices guidelines, *Harv Rev Psychiatry*, (2002) 10(3): 138-153.

[4]   B. Kaplan, Evaluating informatics applications--clinical decision support systems literature review, *Int J Med Inform*, (2001) 64(1): 15-37.

[5]   P.R. Orszag and P. Ellis, The challenge of rising health care costs--a view from the Congressional Budget Office, *N Engl J Med*, (2007) 357(18): 1793-1795.

[6]   G.P. Jackson and J.L. Tarpley, How long does it take to train a surgeon? *BMJ*, (2009) 339: b4260-b4260.

[7]   C.C. Bennett and T.W. Doub, Data mining and electronic health records: Selecting optimal clinical treatments in practice, *Proceedings of the 6th International Conference on Data Mining*, (CSREA Press, Las Vegas, Nevada, 2010) 313-318. http://arxiv.org/abs/1112.1668 (Accessed: 11 May 2012).

[8]   J.A. Osheroff, J.M. Teich, B. Middleton, E.B. Steen, A. Wright, and D.E. Detmer. A roadmap for national action on clinical decision support, *J Am Med Inform Assoc*, (2007) 14(2):141–145.

[9]   C.C. Bennett, T.W. Doub, A.D. Bragg, J. Luellen, C. Van Regenmorter, J. Lockman, et al., Data Mining Session-Based Patient Reported Outcomes (PROs) in a Mental Health Setting: Toward Data-Driven Clinical Decision Support and Personalized Treatment, *IEEE Health Informatics, Imaging, and Systems Biology Conference*, (IEEE, San Jose, CA, 2011) 229-236. http://arxiv.org/abs/1112.1670 (Accessed: 11 May 2012).

[10]  I.S. Kohane, The twin questions of personalized medicine: who are you and whom do you most resemble? *Genome Med*, (2009) 1(1): 4.

[11] Y. Sun Y, S. Goodison S, J. Li, L. Liu, and W. Farmerie, Improved breast cancer prognosis through the combination of clinical and genetic markers, *Bioinformatics*, (2007) 23(1): 30-37.

[12] O. Gevaert, F. De Smet, D. Timmerman, Y. Moreau, and B, De Moor, Predicting the prognosis of breast cancer by integrating clinical and microarray data with Bayesian networks, *Bioinformatics*, (2006) 22(14):184-190.

[13] A.L. Boulesteix, C. Porzelius, and M. Daumer, Microarray-based classification and clinical predictors: on combined classifiers and additional predictive value, *Bioinformatics*, (2008) 24(15): 1698-1706.

[14] J.R. Beck and S.G. Pauker, The Markov process in medical prognosis, *Med Decis Making*, (1983) 3(4):419–58.

[15] Y. Xiang and K.L. Poh, Time-critical dynamic decision modeling in medicine, *Comput Biol Med*, (2002) 32(2): 85-97.

[16] T.Y. Leong, Dynamic decision modeling in medicine: a critique of existing formalisms., *Proc Annu Symp Comput Appl Med Care*, (AMIA, Washington, DC,1993) 478–484.

[17] J.E. Stahl, Modelling methods for pharmacoeconomics and health technology assessment: an overview and guide, *Pharmacoeconomics*, (2008) 26(2): 131–48.

[18] A.J. Schaefer, M.D. Bailey, S.M. Shechter, and M.S. Roberts, Modeling Medical Treatment Using Markov Decision Processes, in: M.L. Brandeau, F. Sainfort, and W.P. Pierskalla, eds., *Operations Research and Health Care*, (Kluwer Academic Publishers, Boston, 2005) 593–612.

[19] O. Alagoz, H. Hsu, A.J. Schaefer, and M.S. Roberts, Markov Decision Processes: A Tool for Sequential Decision Making under Uncertainty, *Med Decis Making*, (2010) 30(4): 474–83.

[20] S.M. Shechter, M.D. Bailey, A.J. Schaefer, and M.S. Roberts, The Optimal Time to Initiate HIV Therapy Under Ordered Health States, *Oper Res*, (2008) 56(1): 20–33.

[21] P.E. Meehl, Causes and effects of my disturbing little book, Journal of Personality Assessment, (1986) 50(3): 370–375.

[22] V.L. Patel, J.F. Arocha, and D.R. Kaufman, A Primer on Aspects of Cognition for Medical Informatics, *J Am Med Inform Assoc*, (2001) 8(4): 324–43.

[23] A.S. Elstein and A. Schwarz, Clinical problem solving and diagnostic decision making: selective review of the cognitive literature, *BMJ*, (2002) 324(7339): 729–32.

[24] M.L. Littman, A tutorial on partially observable Markov decision processes, *J Math Psychol*, (2009) 53(3): 119–25.

[25] S. Russell and P. Norvig, *Artificial Intelligence: A Modern Approach, 3$^{rd}$ Ed*, (Prentice Hall, Upper Saddle River, NJ, 2010).

[26] Y. Gocgun, B.W. Bresnahan, A. Ghate, and M.L. Gunn, A Markov decision process approach to multi-category patient scheduling in a diagnostic facility, *Artif Intell Med*, (2011) 53(2): 73–81.

[27] F. Bousquet and C. Le Page, Multi-agent simulations and ecosystem management: a review, *Ecol Modell*, (2004) 176(3-4): 313–32.

[28] S.D. Miller, B.L. Duncan, J. Brown, R. Sorrell, and M.B. Chalk, Using formal client feedback to improve retention and outcome: Making ongoing, real-time assessment feasible, *Journal of Brief Therapy*, (2006) 5(1): 5–22.

[29] A. Campbell and S. Hemsley, Outcome Rating Scale and Session Rating Scale in psychological practice: Clinical utility of ultra-brief measures, *Clinical Psychologist*, (2009) 13(1): 1–9.

[30] C.C. Bennett, Clinical Productivity System: A Decision Support Model, *International Journal of Productivity and Performance Management*, (2010) 60(3): 311-319.

[31] J. Kreke, M.D. Bailey, A.J. Schaefer, D. Angus, and M.S. Roberts, Modeling hospital discharge policies for patients with pneumonia-related sepsis, *IIE Transactions*, (2008) 40(9): 853–860.

[32] J.E. Goulionis and A. Vozikis, Medical decision making for patients with Parkinson disease under average cost criterion, *Aust New Zealand Health Policy*, (2009) 6: 15.

[33] S. Ross, J. Pineau, S. Paquet, and B. Chaib-draa, Online planning algorithms for POMDPs. *J Artif Intell Res*, (2008) 32:663–704.

[34] M. Hauskrecht and H. Fraser, Planning treatment of ischemic heart disease with partially observable Markov decision processes, *Artif Intell Med*, (2000) 18(3): 221–44.

[35] T. Hester and P. Stone, An Empirical Comparison of Abstraction in Models of Markov Decision Processes, *Proceedings of the ICML/UAI/COLT Workshop on Abstraction in Reinforcement Learning*, (ICML, Montreal, Canada, 2009) 18–23.

[36]  M. Kim, A. Ghate, and M.H. Phillips, A Markov decision process approach to temporal modulation of dose fractions in radiation therapy planning, *Phys Med Biol*, (2009) 54(14): 4455–4476.

[37]  M. Brown, E. Bowring, S. Epstein, M. Jhaveri, R. Maheswaran, P. Mallick, et al, Applying Multi-Agent Techniques to Cancer Modeling, *Proceedings of the 6th Annual Workshop on Multiagent Sequential Decision Making in Uncertain Domains (MSDM)*, (AAMAS, Taipei, Taiwan, 2011) 8–15.

[38]  M.B. Rosenthal, Beyond Pay for Performance -- Emerging Models of Provider-Payment Reform, *N Engl J Med*, (2008) 359(12): 1197-1200.

[39]  Institute of Medicine, *Informing the Future: Critical Issues in Health, 6th Edition*, (The National Academies Press, Washington, D.C., 2011).